%% file: main.tex
\documentclass[runningheads]{llncs}
\usepackage[T1]{fontenc}
\usepackage{graphicx}
\usepackage{booktabs}
\usepackage{multirow}
\usepackage{listings}
\usepackage{amsmath}
\usepackage{xcolor}
\usepackage{enumitem}
\usepackage{hyperref}

\newcommand{\mathcog}{\textsc{MathCog}}
\newcommand\variable[1]{\textcolor{blue}{#1}}
\newcommand\generated[1]{\textcolor{orange}{#1}}

\begin{document}
\title{Benchmarking Large Language Models for Diagnosing Students’ Cognitive Skills from Handwritten Math Work}
\titlerunning{LLMs for Diagnosing Cognitive Skill in Math}
%
%
\authorrunning{Y. Kim et al.}
%
%

\author{Yoonsu Kim\inst{1} \and 
Hyoungwook Jin\inst{2} \and
Hayeon Doh\inst{3} \and
Eunhye Kim\inst{1} \and
Dongyun Jung\inst{1} \and
Seungju Kim\inst{1} \and
Kiyoon Choi\inst{4} \and
Jinho Son\inst{4} \and
Juho Kim\inst{1} \\
}

\institute{School of Computing, KAIST, Daejeon, Republic of Korea
\email{\{yoonsu16,gracekim027,djung2023,sjkim64891,juhokim\}@kaist.ac.kr} \and
University of Michigan, Ann Arbor, USA
\email{jinhw@umich.edu} \and
Ewha Womans University, Seoul, Republic of Korea
\email{hydoh12@ewha.ac.kr} \and
AlgorithmLabs, Seoul, Republic of Korea
\email{\{chlrldbs1,sjhfam\}@algorithmlabs.co.kr}
}

\maketitle              
\begin{abstract}
\input{sections/00_abstract}
\end{abstract}
\input{sections/10_introduction}
\input{sections/20_related_work}
\input{sections/30_dataset}

\input{sections/40_experiment}
\input{sections/50_results}
\input{sections/60_discussion}
\input{sections/70_limitations}
\bibliographystyle{splncs04}
\bibliography{aied2026}
%




\clearpage
\input{sections/80_appendix}

\end{document}

%% file: sections/00_abstract.tex
Students' handwritten math work provides a rich resource for diagnosing cognitive skills, as it captures intermediate reasoning beyond final answers. We investigate how current large language models (LLMs) perform in diagnosing cognitive skills from such work. However, student responses vary widely, often omitting steps or providing only vague, contextually implicit evidence. Despite recent advances in LLMs' multimodal and reasoning capabilities, their performance under such conditions remains underexplored. To address this gap, we constructed \mathcog, a benchmark dataset containing 3,036 diagnostic verdicts across 639 student responses to 110 math problems, annotated by teachers using TIMSS-grounded cognitive skill checklists with evidential strength labels (\textit{Evident}/\textit{Vague}). Evaluating 18 LLMs, we find that (1) all models underperform (F1 < 0.5) regardless of capability, and (2) performance degrades sharply under vague evidence. Error analysis reveals systematic patterns: models frequently misattribute \textit{vague} evidence as \textit{evident}, overthink minimal cues, and hallucinate nonexistent evidence. We discuss implications for evidence-aware, teacher-in-the-loop designs for LLM-based cognitive diagnosis in educational settings. 

%% file: sections/10_introduction.tex
\section{Introduction}
Diagnosing students' cognitive skills from their problem-solving work is a long-standing goal in mathematics education, as it can reveal where and how students' reasoning breaks down beyond final correctness~\cite{livingston2009constructed,jin2024using}. As defined in the TIMSS assessment framework~\cite{mullis2017timss}, \textit{cognitive skills} refer to students' abilities to \textit{know}, \textit{apply}, and \textit{reason} with mathematical concepts and procedures. Handwritten math work is particularly valuable for diagnosing these skills because it can reflect intermediate reasoning steps and partial understanding~\cite{kheong1994information,jin2024using}. 

However, such diagnostic evidence is often incomplete, implicit, or unevenly expressed across students~\cite{henderson2004grading,rahbarnia2014study}, making reliable cognitive skill diagnosis inherently challenging. Despite recent advances in large language models (LLMs), including multimodal perception~\cite{zhang2024mathverse,zhang2024mm} and reasoning capabilities~\cite{wang2023can,hao2024llm}, their ability to diagnose students’ cognitive skills from handwritten work remains largely unexamined. While prior work has examined LLMs' own mathematical problem-solving abilities~\cite{didolkar2024metacognitive,fang2024mathodyssey,ahn2024large}, our work extends beyond problem solving to examine their ability to diagnose human problem-solving processes. 

In this work, we systematically investigate how well existing LLMs diagnose students' cognitive skills in mathematics. Specifically, we focus on the degree of evidential strength in students' handwritten responses. We address the following research questions:
\begin{itemize}[leftmargin=*]
    \item [\textbf{RQ1.}] How do different LLMs (varying in image input, reasoning, model size, and few-shot prompting) perform in diagnosing students' cognitive skills?
    \item[\textbf{RQ2.}] How does the evidential strength of student responses affect LLM's cognitive skill diagnosis performance?
\end{itemize}

To answer these questions, we constructed \mathcog, an expert-crafted benchmark dataset designed to evaluate cognitive skill diagnosis. In collaboration with 5 education experts and 15 middle school teachers, we curated 12 middle school math topics and 110 problems, each with 50+ student responses diagnosed by teachers based on a problem-specific diagnostic checklist grounded in the TIMSS cognitive framework~\cite{mullis2017timss}. Teachers provided binary judgements (\textit{Yes}/\textit{No}) for each skill and annotated the presence of supporting evidence as \textit{Evident} or \textit{Vague}, yielding 3,036 diagnostic items total. Critically, we curated our dataset to include only problems where human experts achieve >70\% agreement, establishing a stable ground truth rather than subjective noise. 

Using \mathcog, we evaluated 18 closed- and open-source LLMs spanning multiple model families, sizes, and capabilities. Results indicate that current LLMs struggle to reliably diagnose students' cognitive skills (all F1 scores < 0.5), with performance degrading sharply under vague-evidence conditions and frequently misattributing weak evidence as strong.
Qualitative error analysis further reveals the model's failure patterns across model types. In particular, models often \textit{misidentify} or \textit{hallucinate} supporting evidence, or \textit{over-infer} students' cognitive skills from incomplete work, showing that errors cascade throughout the diagnostic pipeline from recognizing student responses to interpreting them based on the rubric. Building on these findings, we discuss implications for designing LLM-based diagnostic tools that better elicit students' diagnostic evidence of cognitive processes and support teacher-in-the-loop interpretation.

Our contributions are threefold. (1) We introduce \mathcog, an expert-crafted benchmark dataset for cognitive skill diagnosis. (2) Using \mathcog, we evaluate 18 large language models and analyze the effects of multimodality, reasoning, model size, and evidential strength on diagnosis performance. (3) We analyze diagnostic error patterns under vague-evidence conditions and discuss implications for the design and use of LLM-based cognitive diagnostic systems.

%% file: sections/20_related_work.tex
\section{Related Work}
We review the foundational framework underlying our cognitive diagnosis task, prior investigations into LLMs' performance in math-related contexts, and existing benchmarks, highlighting their current gaps.

\subsection{Cognitive Diagnosis in Mathematics Education}
The TIMSS~\cite{mullis2017timss} is a comprehensive and math-specific framework for cognitive diagnosis, comprising content and cognitive domains. The cognitive domain, which evaluates knowledge application, is divided into three key areas: Knowing (\textit{recalling} definitions, \textit{recognizing} mathematical entities, \textit{classifying or ordering} quantities, \textit{computing}), Applying (\textit{determining} strategies, \textit{representing} problems, \textit{implementing} solution procedures), and Reasoning (\textit{analyzing}, \textit{justifying}).
This research aims to explore whether LLMs can substitute for the mapping to enable scalable and explainable cognitive diagnosis. To our knowledge, this is the first systematic investigation of LLMs in the context of the TIMSS framework.

\subsection{LLM Capabilities in Mathematical Tasks}
Recent advances have enabled LLMs to achieve remarkable performance in mathematical problem-solving~\cite{guo2025deepseek,openai2024}, yet they show relative weaknesses in diagnosing student abilities~\cite{macina2025mathtutorbench,weitekamp2025tutorgym}. These diagnostic challenges are further complicated by current models' struggles with multimodal inputs such as visual elements and handwritten content~\cite{zhang2024mathverse,baral2025,liu2024ai}. While some recent work has demonstrated LLMs' potential for cross-domain cognitive diagnosis~\cite{ma2025large}, existing research has limited systematic exploration of LLMs' diagnostic capabilities for cognitive skills in math specifically. We address this critical gap in the field by providing the first comprehensive evaluation of LLMs' ability to diagnose cognitive skills in mathematical contexts using an established educational framework.

\subsection{Benchmarks for Evaluating Mathematical Assessment Tasks}
In educational settings, several benchmarks have been proposed to evaluate LLMs’ performance in mathematical assessment tasks, including grading, skill recognition, and error analysis. 
For example, MathFish~\cite{lucy2024} aligns 9,900 problems with 385 K–12 standards to assess models’ ability to identify targeted mathematical skills and concepts. Automated scoring of open-ended constructed responses in math word problems has also been studied~\cite{hellman2023}. Other benchmarks focus on misconception detection in multiple-choice questions~\cite{KaggleEedi2020} or handwriting recognition in student work~\cite{baral2025}. While prior research has focused on error detection and scoring, diagnosing complex cognitive skills from students' open-ended handwritten responses remains underexplored and under-resourced. To address this gap, we introduce a novel dataset for cognitive skill diagnosis that enables systematic evaluation of LLMs’ ability to interpret implicit reasoning and diagnose students’ cognitive skills beyond final answers.

%% file: sections/30_dataset.tex
\section{Dataset: \mathcog}
\input{figures/dataset_meta_info}
To evaluate LLMs' performance on cognitive skill diagnosis, we created a benchmark dataset comprising secondary-school math problems, handwritten student responses, diagnostic checklists, and teacher-generated verdicts (Table~\ref{table:dataset_meta_info}). 
The math problem and student response data are from AI-Hub\footnote{The data are available with approval from the provider via \href{https://aihub.or.kr/aihubdata/data/view.do?dataSetSn=71716}{AIHub}.}
, which provides OCR transcriptions of handwritten work. Each topic includes multiple isomorphic problems that share the same problem-solving procedures but differ in numerical values (e.g., ``Solve $x^2+2x-3=0$'' and ``Solve $x^2+x-2=0$'')~\cite{reed1990selecting,morrison2015subgoals}. From this data, we focused on topics from grades 7–9, where problems are sufficiently complex to elicit students’ cognitive processes. We further excluded topics with fewer than 50 student responses to ensure sufficient data for reliable evaluation. We decided not to augment our dataset with synthesized data, as it might bias our observations on this novel task. Through this filtering process, the resulting dataset contains topics, 137 problems, and 796 student responses.

\subsection{Diagnostic Checklist}
For each topic, we developed a diagnostic checklist commonly applicable to isomorphic problems. Each checklist consists of binary question items mapped to one of the 15 cognitive skills defined in the TIMSS 2019 assessment framework~\cite{mullis2017timss}. The checklist items were adapted from TIMSS skill descriptions and refined through expert review by five mathematics curriculum and evaluation experts with PhD degrees in education and practical experience in making math assessment guidelines. Experts gave feedback on the clarity, granularity, and validity of the checklist items.
The experts also commented on the skills each problem can or cannot assess. Experts pointed out that our math problems primarily focus on calculating numbers and applying knowledge, and hence are limited in assessing ``reasoning'' (e.g., justifying, analyzing, generalizing)~\cite{mullis2017timss} by design. We scoped our check items to ``knowing'' and ``applying'' cognitive domains only. We took two iterations to refine the checklists, and each checklist was reviewed by two experts independently in each iteration.

\subsection{Teacher-generated Verdict}
We recruited 15 middle school math teachers to evaluate 796 student responses based on predefined diagnostic checklists. Teachers had an average of 6.1±4.3 years of experience (range: 2.5–20 years). Each check item was assessed along two dimensions: correctness (\textit{Yes}/\textit{No}) and evidential strength (\textit{Evident}/\textit{Vague}). A ``\textit{Yes}'' response indicated that the student fully demonstrated the cognitive action specified, while a ``\textit{No}'' indicated otherwise, including partially demonstrated responses. ``\textit{Evident}'' meant there was clear evidence to support the judgment, whereas ``\textit{Vague}'' signified insufficient evidence. 
\begin{figure}[ht]
\centering
\includegraphics[width=0.9\columnwidth]{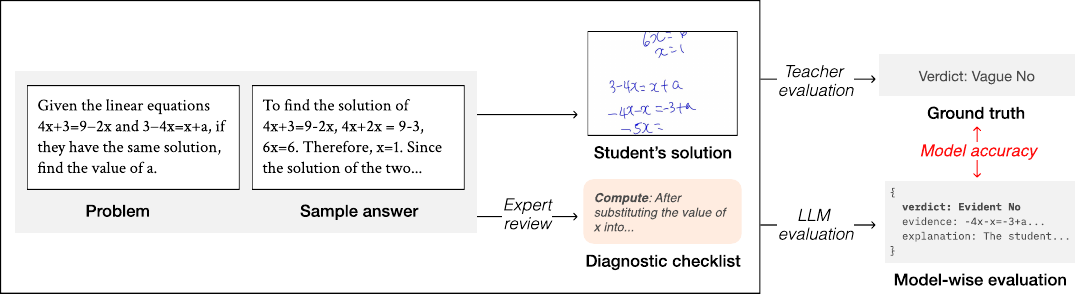}
\caption{Overview of the diagnostic pipeline of the \mathcog.}
\label{fig:conceptual_framework}
\end{figure}
To account for the subjectivity in cognitive diagnosis, each student's response was evaluated by three teachers with overlapping assignments, allowing us to measure inter-rater agreement (see Table ~\ref{table:inter_rater_agreement} in Appendix\footnote{Appendix can be found: \href{https://osf.io/67kz9/overview?view_only=430e88aa45cf493db06f573dfb3eaae2}{OSF appendix materials}}). Following the threshold established in prior literature~\cite{graham2012measuring}, we excluded topics with agreement below 70\%, resulting in a final dataset of 12 topics, 110 problems, and 639 student responses. As a result, the benchmark focuses on diagnostic cases with stable expert agreement, providing a reliable basis for evaluating model performance. These topics cover three-fourths of the content domains defined in TIMSS 2019 (see Table~\ref{table:topics} in Appendix), and the dataset remains reasonably sized given the substantial human effort required for expert annotation.

%% file: figures/dataset_meta_info.tex
\begin{table*}[!htb]
\centering
\scriptsize
\resizebox{\textwidth}{!}{%
\begin{tabular}{llll}
\toprule
\textbf{Problem} &
  \textbf{Student Response} &
  \textbf{Diagnostic Checklist} &
  \textbf{Verdict} \\ \midrule
\multirow{3}{*}{\begin{tabular}[c]{@{}l@{}}When $P=\sqrt{3}$\\ $-7\sqrt{5}+2\sqrt{3}$,\\  $Q=2\sqrt{3}-\sqrt{5}$\\ $+2\sqrt{5}$, find the \\ value of $P+Q$.\end{tabular}} &
  \multirow{3}{*}{\includegraphics[width=0.2\textwidth]{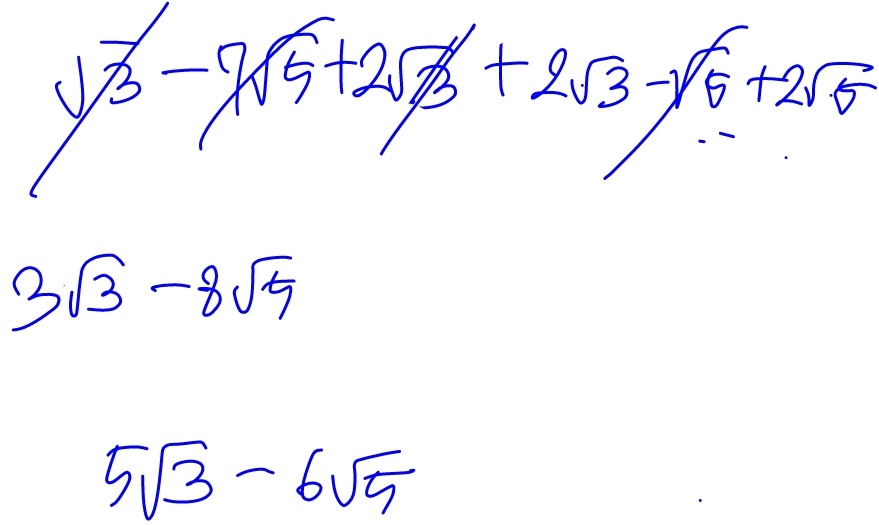}} &
  \begin{tabular}[c]{@{}l@{}}\textbf{Recognize}: Does the student \\ realize that the coefficients can \\ be combined when the radicals \\ are the same?\end{tabular} &
  \begin{tabular}[c]{@{}l@{}}Vague\\ Yes\end{tabular} \\ \cline{3-4} 
 &
   &
  \begin{tabular}[c]{@{}l@{}}\textbf{Compute}: Were the addition \\ and subtraction calculations \\ of coefficients performed \\ accurately?\end{tabular} &
  \begin{tabular}[c]{@{}l@{}}Evident\\ Yes\end{tabular} \\ \cline{3-4} 
 &
   &
  \begin{tabular}[c]{@{}l@{}}\textbf{Determine}: Does the student\\ choose a strategy to first \\ simply organize each equation\\ and then find the final sum?\end{tabular} &
  \begin{tabular}[c]{@{}l@{}}Evident\\ No\end{tabular} \\ \midrule
\multirow{5}{*}{\begin{tabular}[c]{@{}l@{}}There is a trapezoid \\ with a lower side\\ length of $8$ $cm$ \\ and a height of $4$ \\ $cm$. If the area of \\ this trapezoid is not \\ less than $28$ $cm^2$, \\ find how much \\ more $cm$ the length\\ of the upper side of \\ the trapezoid must be.\end{tabular}} &
  \multirow{5}{*}{\includegraphics[width=0.2\textwidth]{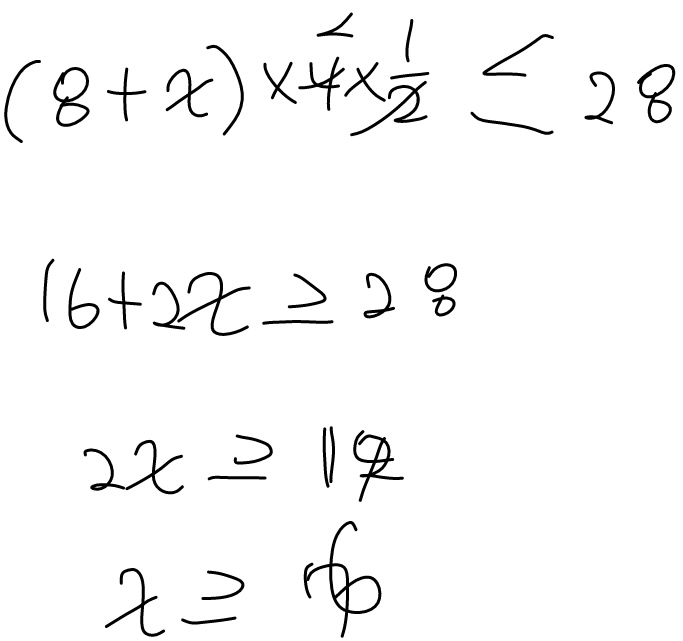}} &
  \begin{tabular}[c]{@{}l@{}}\textbf{Recall}: Does the student \\ remember the formula for \\ the area of a shape correctly? \end{tabular} &
  \begin{tabular}[c]{@{}l@{}}Evident\\ Yes\end{tabular} \\ \cline{3-4} 
 &
   &
  \begin{tabular}[c]{@{}l@{}}\textbf{Compute}: Has the student \\ calculated the linear and \\ constant terms correctly?\end{tabular} &
  \begin{tabular}[c]{@{}l@{}}Vague\\ No\end{tabular} \\ \cline{3-4} 
 &
   &
  \begin{tabular}[c]{@{}l@{}}\textbf{Determine}: Did the student\\ know the need to set up an \\ equation and then solve it \\ to find the range of solutions\\ that meet the conditions?\end{tabular} &
  \begin{tabular}[c]{@{}l@{}}Evident\\ Yes\end{tabular} \\ \cline{3-4} 
 &
   &
  \begin{tabular}[c]{@{}l@{}}\textbf{Represent}: Has the given \\ situation been expressed \\ correctly?\end{tabular} &
  \begin{tabular}[c]{@{}l@{}}Evident\\ No\end{tabular} \\ \cline{3-4} 
 &
   &
  \begin{tabular}[c]{@{}l@{}}\textbf{Implement}: When simplifying\\ an expression, does the student\\ keep the expression correct by\\ performing the same operation\\ on both sides?\end{tabular} &
  \begin{tabular}[c]{@{}l@{}}Vague\\ Yes\end{tabular} \\
  \bottomrule
\end{tabular}
}
\caption{Two samples from \mathcog. Each data point is composed of a math problem, student response, relevant diagnostic checklist, and verdict for each check item.}
\label{table:dataset_meta_info}
\end{table*}

%% file: sections/40_experiment.tex
\section{Experimental Setting}
Using \mathcog~as a benchmark dataset, we evaluated a diverse set of LLMs with varying input modalities, reasoning capabilities, model sizes, and prompting strategies to address RQ1, and analyzed how the evidential strength of student responses affects diagnosis performance to address RQ2.



\textit{\textbf{Prompting.}}
We instructed LLMs to evaluate each diagnostic check item given a math problem, its solution, a student’s response, and a diagnostic checklist. Student's response image inputs were provided as OCR transcriptions, with mathematical formulas and visual cues (e.g., strikethroughs) represented in LaTeX. For multimodal settings, we additionally supplied images of student responses. Since the original inputs were in Korean, we machine-translated them into English to prevent possible performance degradation due to language~\cite{achiam2023gpt}; we used Google Translate API for batch translation and manually verified them. We employed Chain-of-Thought prompting~\cite{liu2024ai} to guide LLMs to systematically address each check item by first restating its content, identifying relevant evidence, providing an explanation, and delivering a final verdict. The verdict followed one of the four categories used by teachers in \mathcog. 
To examine the effect of in-context examples on model performance, we experimented with few-shot prompting using two exemplars that covered all four types of verdicts. To minimize randomness, all LLMs were run with a temperature setting of zero. Full system and user prompts are provided in the Appendix.

\textit{\textbf{Models.}}
We evaluated 18 LLMs from multiple model families, selected to analyze the effects of multimodality, reasoning capability, and model size. For multimodality, selected models were tested under both text-only and image-augmented conditions; for reasoning, we compared reasoning-oriented models with their conventional counterparts within the same model families.

\textit{\textbf{Metrics.}}
We evaluated LLM outputs against teacher-provided ground-truth diagnostic labels. Overall diagnosis performance was assessed using \textbf{macro F1 score} and \textbf{accuracy} over the four verdict categories.
We report macro F1 to mitigate the effect of label imbalance across verdicts (see Table~\ref{table:distribution} in Appendix).
To address \textbf{RQ2}, we analyzed how the \textit{evidential strength} of student responses affects diagnosis performance. To formalize this analysis, we treat each diagnosis instance as a $(\textit{student response}, \textit{cognitive skill})$ pair, indexed by $i$. For each instance $i$, the ground-truth labels consist of a skill judgment $y_i \in \{\text{Yes}, \text{No}\}$ and evidential strength $e_i \in \{\text{Evident}, \text{Vague}\}$, and the corresponding model predictions $\hat{y}_i$ and $\hat{e}_i$.
To characterize how models handle evidential strength, we define two complementary metrics that capture distinct failure modes in evidence attribution. 
First, we define \textbf{evidence over-attribution} (\textbf{OverAttr}) as the frequency with which a model assigns ``Evident'' to cases where the ground-truth evidential strength is ``Vague'': 
\begin{equation}
\text{OverAttr} =\frac{\left|\left\{ i \mid e_i = \text{Vague} \wedge \hat{e}_i = \text{Evident} \right\}\right|}{\left|\left\{ i \mid e_i = \text{Vague} \right\}\right|}
\end{equation}
This metric reflects the model's tendency to overstate evidential support under vague evidence cases, independent of diagnosis correctness. 
Second, we define \textbf{evidence false-attribution} (\textbf{FalseAttr}) as the frequency with which a model assigns Evident to incorrect diagnoses: 
\begin{equation}
\text{FalseAttr} =\frac{\left|\left\{ i \mid \hat{y}_i \neq y_i \wedge \hat{e}_i = \text{Evident} \right\}\right|}{\left|\left\{ i \mid \hat{y}_i \neq y_i \right\}\right|}
\end{equation}
This metric captures a particularly misleading failure mode, where erroneous diagnoses are accompanied by strong evidential claims.

%% file: sections/50_results.tex
\section{Results}
This section reports results addressing our research questions.
\input{tables/result_all}

\begin{figure}[ht]
\centering
\includegraphics[width=\linewidth]{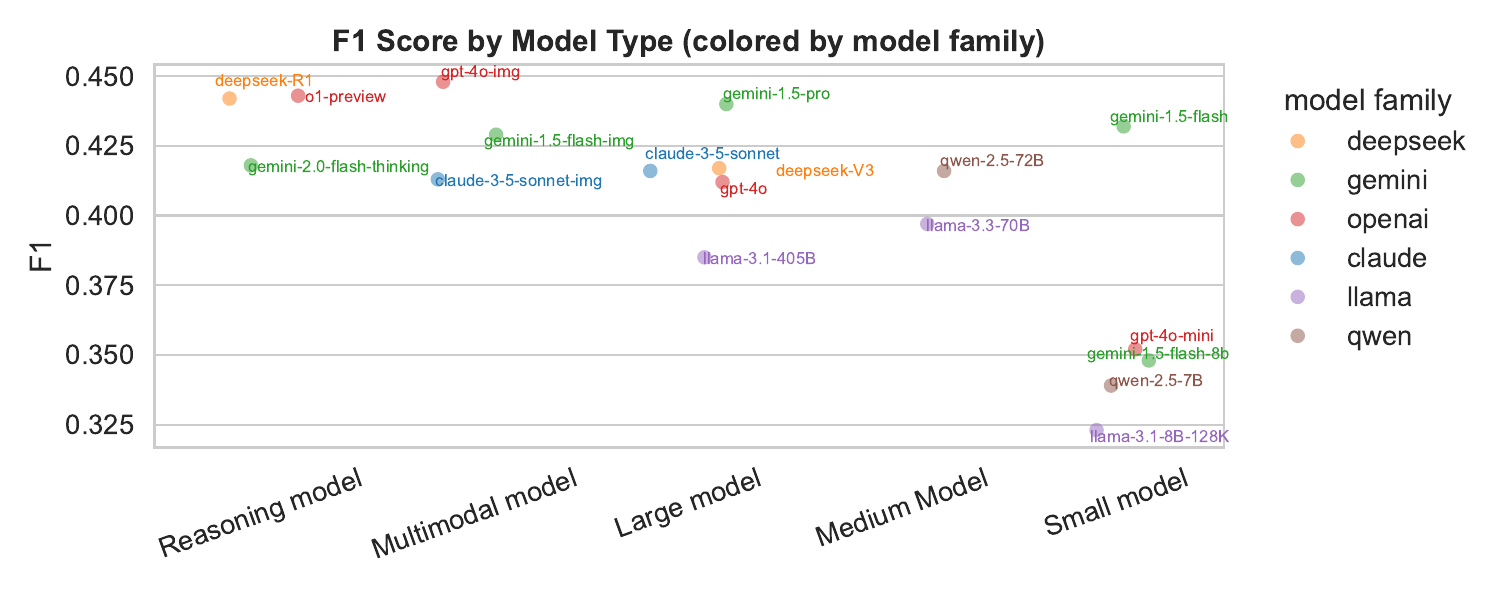}
\caption{F1 score across by model type and family.}
\label{fig:f1_model_type}
\end{figure}

\subsection{RQ1. How do different LLMs perform in diagnosing students' cognitive skills?}
Overall, all evaluated LLMs showed limited performance in cognitive skill diagnosis, with macro F1 scores below 0.5 (Figure~\ref{fig:f1_model_type}, Table~\ref{table:result_all}).
While overall accuracy was relatively moderate ($M=.680$, $SD=.050$), it obscures important diagnostic failures. Models frequently over-attributed evidential strength (OverAttr; $M=.580$, $SD=.120$) and falsely attributed evidence (FalseAttr; $M=.585$, $SD=.145$), often asserting strong evidence even when diagnoses were incorrect. Notably, even the best-performing models in terms of F1 score and accuracy (e.g., \texttt{DeepSeek-R1}, \texttt{GPT-4o-img}) showed high OverAttr and FalseAttr values, suggesting that performance gains often co-occur with unreliable evidential judgement.

\begin{figure*}[h]
\centering
\includegraphics[width=\linewidth]{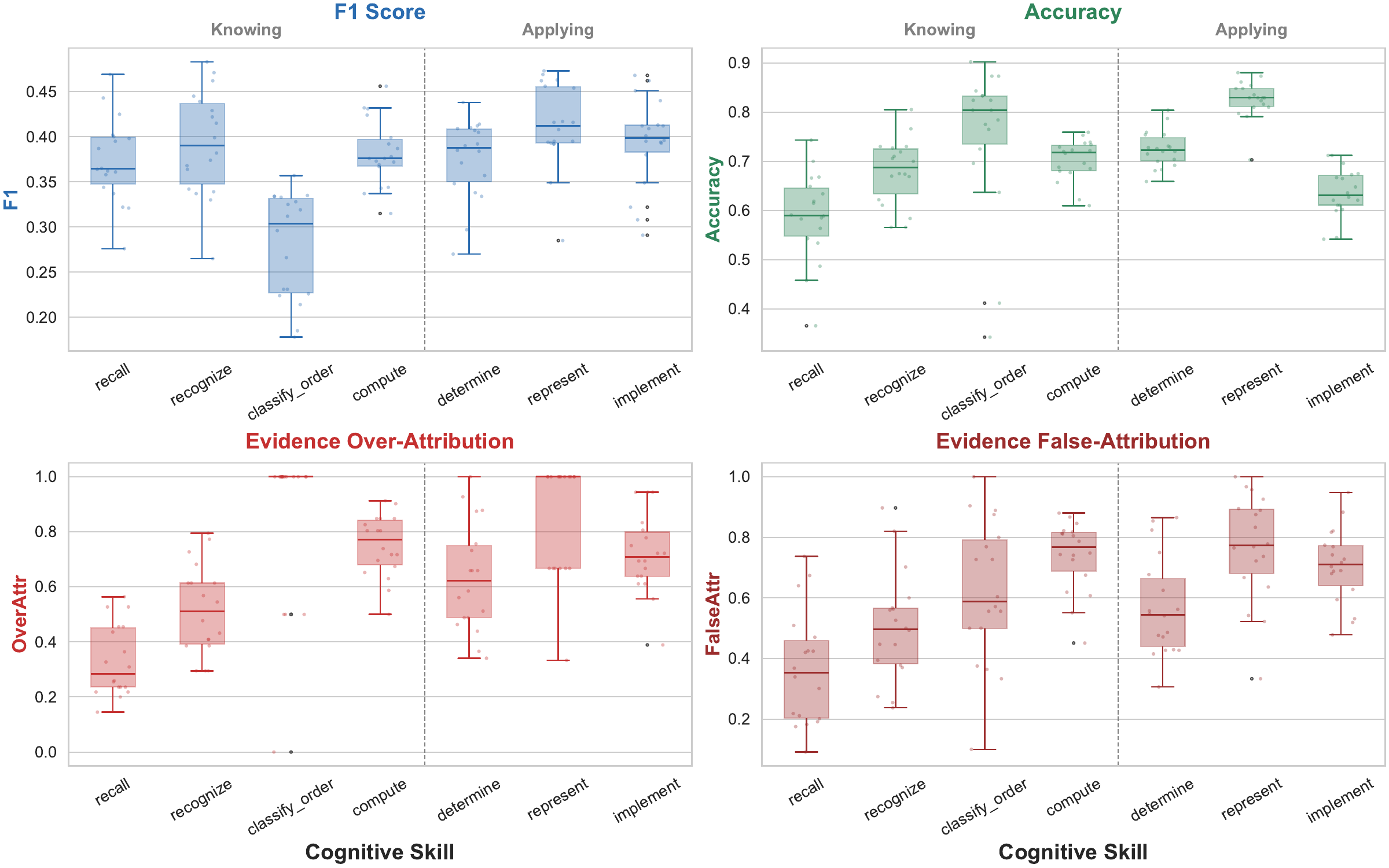}
\caption{Performance and evidential attribution errors across cognitive skills, measured by F1 score, accuracy, evidence over-attribution, and false-attribution.}
\label{fig:result_skills} 
\end{figure*}
Analysis of skill-specific performance (Figure~\ref{fig:result_skills}) reveals further insights into the limitations of current LLMs. \textbf{No skill category achieved an F1 score above 0.5}, indicating that fine-grained diagnosis remains challenging even at the individual skill level. Despite this overall limitation, we observed a performance gap between the \textit{Knowing} and \textit{Applying} cognitive skills. Contrary to our expectation that \textit{Knowing} skills (e.g., \textit{Recall}, \textit{Recognize}) would be easier due to their surface-level nature, models performed better on \textit{Applying} skills in both F1 score ($t=4.34$, $p<.001$) and accuracy ($t=4.85$, $p<.001$). However, this performance gain came with increased evidential errors. \textit{Applying} skills exhibited higher evidence over-attribution ($t=5.44$, $p<.001$) and false-attribution ($t=7.66$, $p<.001$) than \textit{Knowing} skills, indicating a tendency to overstate evidential support when diagnosing higher-level cognitive skills.

\input{tables/fewshot}
We further examined the impact of \textbf{multimodality}, \textbf{reasoning}, \textbf{model size}, and \textbf{few-shot prompting} as factors that may influence diagnostic performance (Table ~\ref{table:result_all}).
\textbf{Multimodal input provided modest but inconsistent benefits.} Models supporting image input consistently outperformed their text-only counterparts in accuracy, indicating that access to handwritten layouts and visual cues helps mitigate errors introduced by OCR transcription. However, improvements in F1 score were inconsistent, suggesting that visual input can sometimes introduce noise and does not uniformly improve fine-grained skill diagnosis.
\textbf{Reasoning-oriented models produced similarly mixed results.} While some reasoning-oriented models (e.g., \texttt{DeepSeek-R1}, \texttt{GPT-o1}) achieved higher F1 scores and accuracy than their non-reasoning counterparts, \texttt{Gemini} performed worse. These results suggest that reasoning mechanisms do not uniformly translate into improved cognitive diagnosis. 
\textbf{Model size showed a moderate positive relationship with performance.} Larger models generally achieved higher F1 scores than smaller ones, indicating that increased parameter capacity supports a more robust interpretation of student responses. However, the correlation was moderate (Spearman’s $\rho=.570, p=.053$), suggesting that scale alone is insufficient to account for the substantial variability in performance across models.
Finally, \textbf{few-shot prompting did not improve diagnostic performance.} While in-context examples might seem helpful for such nuanced tasks, F1 and accuracy often decreased rather than improved (Table~\ref{table:fewshot}), suggesting the challenge lies in the complexity of cognitive skill diagnosis rather than prompt design. However, few-shot prompting did reduce tendencies toward evidence over-attribution and false-attribution, indicating that examples may encourage more cautious, evidence-grounded responses even when overall diagnostic accuracy remains limited.
Taken together, these findings suggest that prompt engineering and model capabilities have limited impact on improving the performance of cognitive skill diagnosis. 

\subsection{RQ2. How does the evidential strength of student responses affect LLM's cognitive skill diagnosis performance?}

\begin{figure*}[ht]
\centering
\includegraphics[width=\linewidth]{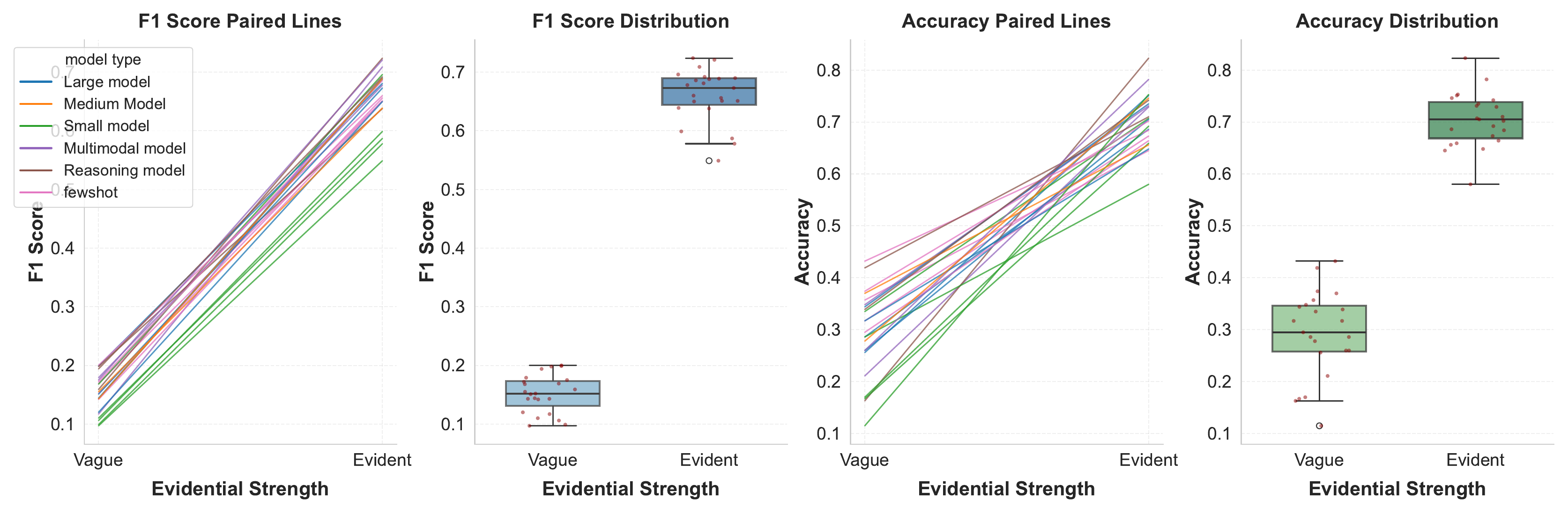}
\caption{Effect of evidential strength on LLM diagnostic performance for every model.}
\label{fig:strong_weak}
\end{figure*}
Our results reveal a \textbf{strong dependency of LLM diagnostic performance on the evidential strength} of student responses. Across all models, both F1 score ($t=64.60$, $p<.001$) and accuracy ($t=17.80$, $p<.001$) were substantially higher when student responses contained \textit{evident} evidence for a cognitive skill than when evidence was \textit{vague} (Fig.~\ref{fig:strong_weak}).
The magnitude of this gap was substantial (mean $\Delta F1 =.51$; mean $\Delta Acc = .41$), indicating that current LLMs rely heavily on explicit evidence and exhibit limited robustness when reasoning under weak or implicit evidential conditions.

Beyond degraded performance, \textbf{models also exhibited over-attribution bias under vague evidence.} Specifically, they frequently labeled responses as \textit{Evident} even when the ground-truth evidential strength was \textit{Vague}, resulting in consistently high \textbf{OverAttr} values.
More concerningly, models also showed a strong tendency toward \textbf{FalseAttr}, in which they asserted \textit{evident} evidence despite arriving at incorrect cognitive skill diagnoses. In such cases, models produced unsupported explanations that appeared plausible yet were not grounded in students’ actual responses.
Notably, these two tendencies were strongly correlated across models (Spearman’s $\rho=.837$, $p<.001$), indicating that models prone to over-attributing evidential strength are also more likely to produce incorrect diagnoses with asserted evidence. This behavior is particularly problematic for educational use, as it can mislead teachers and students by presenting incorrect diagnoses with seemingly well-justified rationales~\cite{kim2025fostering}.

\input{tables/error_code}
To better understand how such errors arise, we conducted a qualitative error analysis of model outputs under incorrectly diagnosed vague evidence cases. We examined all components generated in the models’ CoT responses, including the identified evidence, explanations, and final verdicts, focusing on how the models misinterpreted students' answers, generated unsupported reasoning, and ultimately reached incorrect diagnoses. Three authors collaboratively analyzed 372 cases (13.2\%) sampled from a total of 2,816 error instances, covering 18 model responses, and iteratively derived five recurring error types with substantial inter-rater reliability (Fleiss' $\kappa = 0.74$). The remaining cases were then independently labeled using the agreed error taxonomy. 

\begin{figure*}[ht]
\centering
\includegraphics[width=\linewidth]{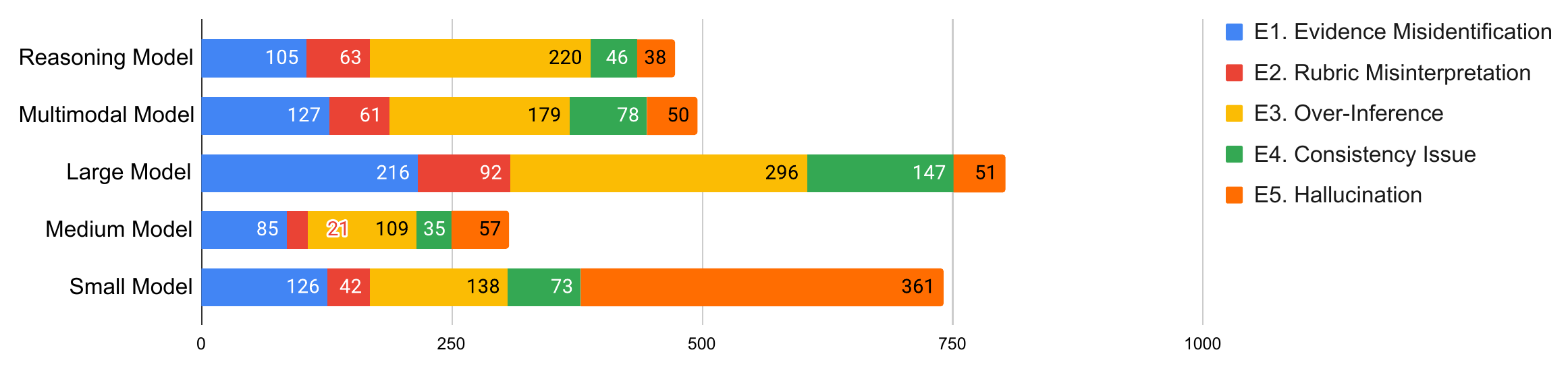}
\caption{Error type distribution by model category.}
\label{fig:error_code_model}
\end{figure*}

Table ~\ref{tab:error_code} summarizes the distribution of error types across model categories. Errors related to \textit{over-inference} were most prevalent (33.5\%), followed by \textit{evidence misidentification} (23.4\%) and \textit{hallucination} (19.8\%). We further observed systematic differences across model types (Figure ~\ref{fig:error_code_model}): smaller models exhibited a higher proportion of \textit{hallucination} errors (48.78\%), whereas reasoning-oriented models committed \textit{over-inferred} more frequently (46.61\%) beyond the information explicitly provided in student responses. 

%% file: tables/result_all.tex
\begin{table*}[ht]
\centering
\resizebox{\textwidth}{!}{%
\begin{tabular}{llcccccccc}
\toprule
\textbf{Category} &
\textbf{Model} &
\multicolumn{1}{l}{\textbf{Precision}} &
\multicolumn{1}{l}{\textbf{Recall}} &
\multicolumn{1}{l}{\textbf{F1}} &
\multicolumn{1}{l}{\textbf{Accuracy}} &
\multicolumn{1}{l}{\textbf{OverAttr}} &
\multicolumn{1}{l}{\textbf{FalseAttr}} \\
\midrule
Reasoning Model & DeepSeek-R1               & .447 & .467 & .442 & {\color{blue}\textbf{.773}} & {\color{red}\textbf{.767}} & {\color{red}\textbf{.862}} \\
                & Gemini-2.0-Flash-Thinking & {\color{blue}\textbf{.454}} & .486 & .418 & .688 & {\color{blue}\textbf{.361}} & .432 \\
                & o1-Preview                & .423 & {\color{blue}\textbf{.513}} & .443 & .711 & .489 & .650 \\
\midrule
Multimodal Model & Claude-3.5-Sonnet-img    & .408 & .457 & .413 & .691 & .626 & .559 \\
                 & Gemini-1.5-Flash-img     & .431 & .484 & .429 & .702 & .471 & .471 \\
                 & GPT-4o-img               & .427 & .494 & {\color{blue}\textbf{.448}} & .743 & .656 & .681  \\
\midrule
Large Model & Claude-3.5-Sonnet             & .411 & .487 & .416 & .656 & .551 & .574 \\
            & DeepSeek-V3                  & .406 & .461 & .417 & .714 & .612 & .595 \\
            & Gemini-1.5-Pro               & .429 & .496 & .440 & .706 & .520 & .635 \\
            & GPT-4o                       & .398 & .476 & .412 & .672 & .648 & .658 \\
            & Llama-3.1-405B               & .415 & .439 & .385 & .624 & .467 & {\color{blue}\textbf{.307}} \\
\midrule
Medium Model & Llama-3.3-70B                & .415 & .467 & .397 & .635 & .436 & .409 \\
             & Qwen-2.5-72B                 & .415 & .459 & .416 & .711 & .568 & .500 \\
\midrule
Small Model & Gemini-1.5-Flash              & .432 & .490 & .432 & .679 & .502 & .465 \\
            & Gemini-1.5-Flash-8b           & .359 & .427 & .348 & {\color{red}\textbf{.558}} & .537 & .525 \\
            & GPT-4o-mini                  & {\color{red}\textbf{.345}} & .398 & .352 & .622 & {\color{red}\textbf{.767}} & .784 \\
            & Llama-3.1-8B-128K             & .352 & {\color{red}\textbf{.334}} & {\color{red}\textbf{.323}} & .653 & .709 & .624 \\
            & Qwen-2.5-7B                  & .347 & .337 & .339 & .705 & .758 & .798 \\

\bottomrule
\end{tabular}
}
\caption{Performance of all 18 LLMs tested on skill diagnosis tasks. For each metric, {\color{blue}blue} and {\color{red}red} fonts indicate the best and worst values, respectively.}
\label{table:result_all}
\end{table*}

%% file: tables/fewshot.tex
\begin{table*}[ht]
\centering
\resizebox{.8\textwidth}{!}{%
\begin{tabular}{lcccccc}
\toprule
\textbf{Few-shots} & \textbf{Precision} & \textbf{Recall} & \textbf{F1} & \textbf{Accuracy} & \textbf{OverAttr} & \textbf{FalseAttr} \\ 
\midrule
claude-3-5-sonnet & .398          & .485          & .410          & .646          & \textbf{.515} & .647          \\
deepseek-V3       & .404          & \textbf{.493} & .404          & .624          & \textbf{.493} & \textbf{.497} \\
gemini-1.5-pro    & .429          & \textbf{.517} & .430          & .665          & \textbf{.427} & \textbf{.552} \\
gpt-4o            & .390          & \textbf{.485} & .396          & .637          & \textbf{.577} & \textbf{.587} \\
llama-3.1-405B    & \textbf{.442} & \textbf{.494} & \textbf{.429} & \textbf{.677} & .476          & .457          \\
\bottomrule
\end{tabular}%
}
\caption{Performance of five state-of-the-art models in few-shot experiments. Bold values denote improvements over the corresponding zero-shot results.}
\label{table:fewshot}
\end{table*}

%% file: tables/error_code.tex
\begin{table}[]
\resizebox{\columnwidth}{!}{%
\begin{tabular}{lll}
\toprule
\textbf{Error code} & \textbf{Description } & \multicolumn{1}{l}{\textbf{Count (\%)}} \\
\midrule
E1. Evidence Misidentification & Incorrectly identifying or relying on insufficient evidence from the student response. & \textbf{659 (23.40\%)} \\
E2. Rubric Misinterpretation   & Misunderstanding the diagnostic rubric or evaluation criteria.        & \textbf{279 (9.91\%)} \\
E3. Over-Inference             & Inferring unstated reasoning beyond the information provided in the student response.  & \textbf{942 (33.45\%)} \\
E4. Consistency Issue          & Inconsistencies between evidence, explanation, and final verdict.     & \textbf{379 (13.46\%)} \\
E5. Hallucination              & Introducing content or reasoning not present in the student response. & \textbf{557 (19.78\%)} \\
\bottomrule
\end{tabular}%
}
\caption{Error types and their distribution identified from qualitative analysis of LLM diagnostic failures under vague evidence.}
\label{tab:error_code}
\end{table}

%% file: sections/60_discussion.tex
\section{Discussion}
Our findings imply that current LLMs are not yet reliable for cognitive skill diagnosis without intermediate verification, as errors cascade throughout the diagnostic pipeline. Performance was strongly constrained by the evidential strength in student responses rather than by model capabilities. This challenge is closely tied to the nature of students' handwritten work, which often omits intermediate reasoning and expresses ideas in implicit or non-linear ways, making diagnostic evidence difficult to assess~\cite{henderson2004grading,rahbarnia2014study}. Under such conditions, LLMs tend to over-attribute evidential strength and, in some cases, arrive at incorrect diagnoses with asserted evidence, raising concerns for real-world educational use.

Our findings point to an important opportunity: richer diagnostic evidence could improve cognitive diagnosis. One promising direction is to design problem-solving tasks and interfaces that better elicit students’ reasoning processes and justifications, help them organize their steps tidily, and prompt reflection on strategy choices. For example, interactive interfaces could track diagnostic evidence in real-time and request clarification when evidence appears weak, shifting from post-hoc inference to active elicitation of cognitive processes. 

Beyond task design, our qualitative error analysis highlights the importance of teacher-in-the-loop approaches for mitigating diagnostic failures. Diagnostic errors followed recurring patterns across key stages of the diagnostic process, including identifying evidence, interpreting diagnostic criteria, and synthesizing intermediate reasoning into final judgments. At these stages, teachers can intervene in targeted ways. Errors such as \textit{evidence misidentification} can be mitigated by having teachers validate whether the identified evidence is actually present in the student’s work. \textit{Rubric misinterpretation} can be addressed by allowing teachers to check and correct how model interpretations align with diagnostic criteria. For \textit{over-inference}, teachers can scrutinize whether inferred cognitive skills are sufficiently supported by observable evidence, preventing unwarranted extrapolation. \textit{Consistency issues} and \textit{hallucinations} require human oversight to ensure coherence between intermediate reasoning and final diagnostic judgments.
Rather than treating model outputs as final assessments, such selective human intervention can reduce diagnostic errors, consistent with prior work on evidence-supported reasoning~\cite{wang2024multi} and human-AI complementary assessment workflows~\cite{strong2025trustworthy}.
Overall, responsibly integrating LLMs into educational assessment requires moving beyond fully automated diagnosis toward transparency and selective human verification. 

%% file: sections/70_limitations.tex
\section{Future Work}
While we present a novel, carefully crafted benchmark for evaluating LLM-based cognitive skill diagnosis, several directions remain for future work.
First, the benchmark can be extended in a more targeted manner. Synthetic data generation~\cite{AnabyTavor2019DoNH} could selectively augment diagnostically challenging cases, such as student responses with implicit or vague evidence. Our analysis of 3,036 verdicts highlights where such challenges arise, motivating controlled data expansion rather than indiscriminate scaling.
Second, future work can broaden the range of cognitive demands represented in the dataset. Reasoning-oriented problems that require students to \textit{generalize} or \textit{justify} their thinking are currently underrepresented in \mathcog, and expanding coverage across problem types, topics, and grade levels would enable more comprehensive evaluation.
Finally, beyond dataset expansion, future work can explore richer evaluation settings for cognitive diagnosis, such as human–AI collaborative diagnosis scenarios or analyzing how LLMs revise diagnostic judgments when provided with additional evidence. It could shed light on the reliability and practical use of LLM-based diagnostic systems in educational contexts.

%% file: sections/80_appendix.tex
\appendix
\section{Appendix}

\subsection{Prompts}
The \variable{blue} text represents the programmatically filled arguments, and the \generated{orange} text represents LLM-generated output.

\subsection{System Prompt}
\input{appendix/system_prompt}

\subsection{User Prompt}
\input{appendix/user_prompt}

\subsection{MathCog Dataset Details}
This section provides supplementary details of the \mathcog{} dataset, including the diagnostic workflow, cognitive skill taxonomy, inter-rater reliability of teachers' diagnoses, dataset composition, and representative failure cases of model diagnoses.


\input{appendix/timss_skills}

\input{appendix/topics}

\input{appendix/inter_rater_agreement}

\input{appendix/distribution}

\begin{figure*}[ht]
\centering
\includegraphics[height=8cm,keepaspectratio]{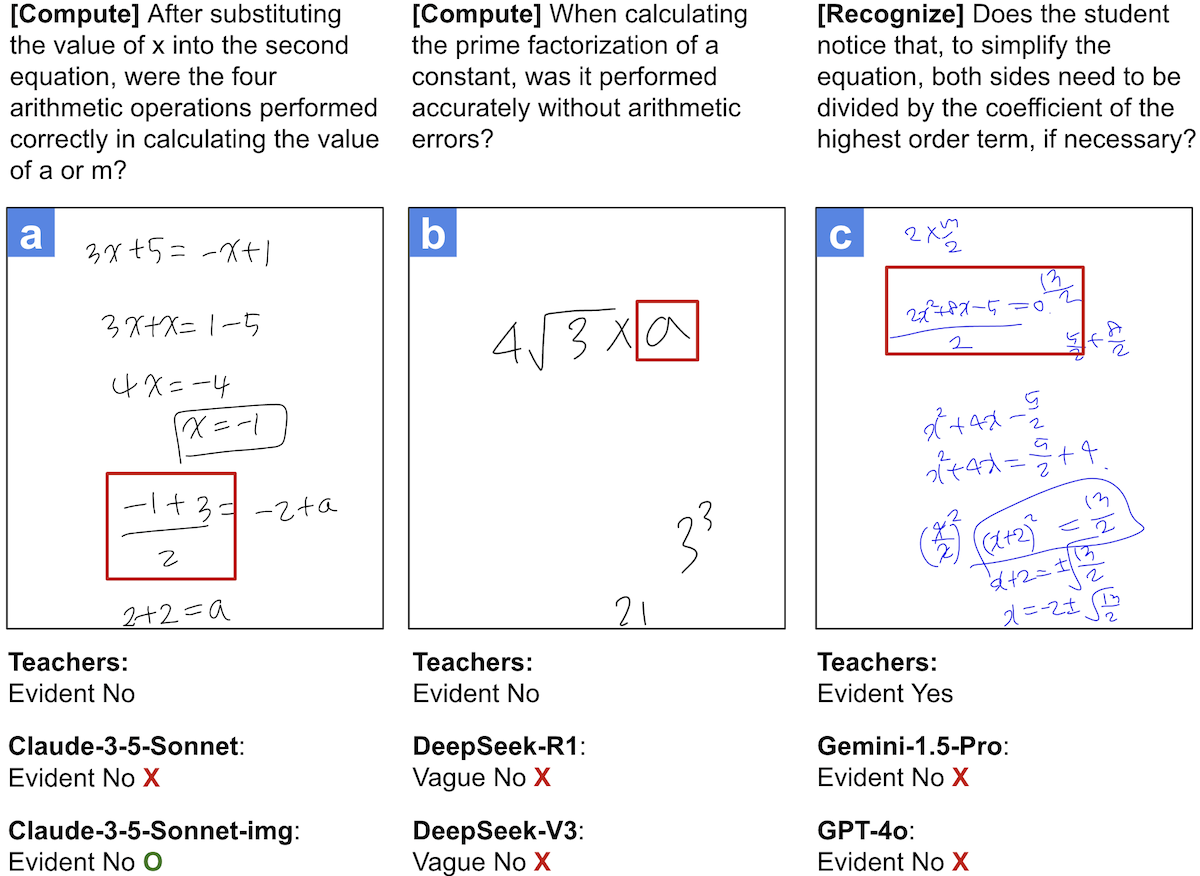}
\caption{Illustrative examples of diagnosis check items and student responses that LLMs failed to diagnose correctly. Evidence for human judgment is marked with a red box.}
\label{fig:fail_cases}
\end{figure*}

%% file: appendix/system_prompt.tex
\begin{lstlisting}[escapechar=\%]
# **Task Description**
You are a middle school math teacher tasked with evaluating students' mathematical thinking skills based on their responses to math problems. Your goal is to analyze a given student's response and determine whether they exhibit specific cognitive skills in solving the problem. Your evaluation must be **strict** and **evidence-based**, meaning that every verdict must be backed by direct evidence from the response. If no clear evidence exists, do not assume correctness.

## **Evaluation Categories**
For each thinking skill in the checklist, you must classify the student's performance into one of the following categories:

- **Evident Yes**: The student's response provides clear and explicit evidence that the check item is met. A direct quote from the response can confirm this.
- **Vague Yes**: The response suggests that the check item might be satisfied, but no specific part of the response directly proves it.
- **Evident No**: The response explicitly contradicts or fails to meet the check item, with clear evidence demonstrating the error or omission.
- **Vague No**: The response does not appear to satisfy the check item, but there is no direct evidence confirming whether the student considered it or not.

## **Input Format**
You will receive the following data:
- **Problem**: A math problem given to a student.
- **Answer**: The correct step-by-step solution.
- **Response**: The student's response to the problem.
- **Check Items**: A set of specific skills to evaluate.

## **Output Format**
Return a valid JSON object structured as follows:
```json
{
    "skills": [
        {
            "checkItem": "<Check Item's [Label] and the Following Question>",
            "evidence": "<Directly Quoted Part of Response>",
            "explanation": "<Explanation About Why the Evidence Supports the Verdict>",
            "verdict": "Evident Yes" | "Vague Yes" | "Evident No" | "Vague No"
        }
    ]
}
```
\end{lstlisting}

%% file: appendix/user_prompt.tex
\begin{lstlisting}[escapechar=\%]
# **Task**
student responses to math problems, extract direct evidence, and strictly classify thinking skills according to the given categories.

Return a valid JSON object structured as follows:
```json
{
    "skills": [
        {
            "checkItem": "<Check Item's [Label] and the Following Question>",
            "evidence": "<Directly Quoted Part of Response>",
            "explanation": "<Explanation About Why the Evidence Supports the Verdict>",
            "verdict": "Evident Yes" | "Vague Yes" | "Evident No" | "Vague No"
        }
    ]
}
```

**Problem:**
%\variable{There is a two-digit natural number whose tens digit is $1$. If the number that changes the tens and ones digits of this natural number is $9$ less than $5$ times the first number, find the first number.}%

**Answer:**
%\variable{If the number in the ones place is $x$, this natural number is $10+x$. The number where the tens digit and the ones digit are swapped is $10x+1$ because the tens digit is $x$ and the ones digit is $1$. The changed number is $9$ smaller than $5$ times the first number, so $10x+1=5(10+x)-9$ $10x+1=5x+41$ $5x=40$ $x=8$ Therefore, the first number is $18$.}%

**Response:**
%\variable{$1$ $x$ $x$ $1$ $10 x+1=5(10+x)-9$ $10 x+1=50+5 x-9$ $41$ $5 x=40$ $x=8$}%

**Check Items:**
%\variable{
- [recognize\textunderscore 0] Is the student aware that the relationship between two numbers switched can be expressed as addition/subtraction/multiplication?}%
%\variable{
- [recognize\textunderscore 1] Has the student figured out all the elements necessary to solve the problem (original numbers, swapped numbers, difference or multiple relationships, etc.)?}%
%\variable{
- [compute\textunderscore 0] Has the student performed algebraic calculations correctly, such as expanding expressions and organizing similar terms?}%
%\variable{
- [determine\textunderscore 0] Has the student chosen the appropriate approach to solving the problem (e.g., establishing a linear equation)?}%
%\variable{
- [represent\textunderscore 0] Are two-digit numbers expressed correctly in the form 10x + y (or equivalent)?}%
%\variable{
- [implement\textunderscore 0] Does the student proceed consistently with the process of expanding the equation and moving terms to find x (or y) according to the planned solution strategy?}%

**Output:**
%\generated{
\{
}%
    %\generated{"skills": [}%
        %\generated{\{}%
            %\generated{
            "checkItem": "[recognize\textunderscore 0] Is the student aware that the relationship between two numbers switched can be expressed as addition/subtraction/multiplication?",}%
            %\generated{
            "evidence": "10 x+1=5(10+x)-9",}%
            %\generated{
            "explanation": "The student sets up the equation 10x+1=5(10+x)-9, which shows they understand the relationship between the original and swapped numbers involves subtraction and multiplication.",}%
            %\generated{
            "verdict": "Evident Yes"}%
        %\generated{\}, \{}%
            %\generated{
            "checkItem": "[recognize\textunderscore 1] Has the student figured out all the elements necessary to solve the problem (original numbers, swapped numbers, difference or multiple relationships, etc.)?",}%
            %\generated{
            "evidence": "10 x+1=5(10+x)-9",}%
            %\generated{
            "explanation": "The student uses the correct equation to represent the relationship between the original and swapped numbers, indicating they have identified the necessary elements.",}%
            %\generated{
            "verdict": "Evident Yes"}%
        %\generated{\},}%
        %\generated{...}%
    %\generated{]}%
%\generated{\}%}
%\generated{```}%
\end{lstlisting}

%% file: appendix/timss_skills.tex
\begin{table*}[ht]
\resizebox{\textwidth}{!}{%
\begin{tabular}{lll}
\toprule
\textbf{Domains} &
  \textbf{Cognitive Skills} &
  \textbf{Description} \\ \midrule
\multirow{6}{*}{Knowing} &
  Recall &
  \begin{tabular}[c]{@{}l@{}}Recall definitions, terminology, number properties, units of measurement, \\ geometric properties, and notation (e.g., $a \times b= ab, a + a + a = 3a$).\end{tabular} \\ \cline{2-3} 
 &
  Recognize &
  \begin{tabular}[c]{@{}l@{}}Recognize numbers, expressions, quantities, and shapes. Recognize entities \\ that are mathematically equivalent (e.g., equivalent familiar fractions, \\ decimals, and percents; different orientations of simple geometric figures).\end{tabular} \\ \cline{2-3} 
 &
  Classify/Order &
  Classify numbers, expressions, quantities, and shapes by common properties. \\ \cline{2-3} 
 &
  Compute &
  \begin{tabular}[c]{@{}l@{}}Carry out algorithmic procedures for $+, -, \times, \div$ or a combination of these \\ with whole numbers, fractions, decimals, and integers. Carry out \\ straightforward algebraic procedures.\end{tabular} \\ \cline{2-3} 
 &
  Retrieve &
  Retrieve information from graphs, tables, texts, or other sources. \\ \cline{2-3} 
 &
  Measure &
  Use measuring instruments; and choose appropriate units of measurement. \\ \midrule
\multirow{3}{*}{Applying} &
  Determine &
  \begin{tabular}[c]{@{}l@{}}Determine efficient/appropriate operations, strategies, and tools for solving \\ problems for which there are commonly used methods of solution.\end{tabular} \\ \cline{2-3} 
 &
  Represent/Model &
  \begin{tabular}[c]{@{}l@{}}Display data in tables or graphs; create equations, inequalities, geometric \\ figures, or diagrams that model problem situations; and generate equivalent \\ representations for a given mathematical entity or relationship.\end{tabular} \\ \cline{2-3} 
 &
  Implement &
  \begin{tabular}[c]{@{}l@{}}Implement strategies and operations to solve problems involving familiar \\ mathematical concepts and procedures.\end{tabular} \\ \midrule
\multirow{6}{*}{Reasoning} &
  Analyze &
  \begin{tabular}[c]{@{}l@{}}Determine, describe, or use relationships among numbers, expressions, \\ quantities, and shapes.\end{tabular} \\ \cline{2-3} 
 &
  \begin{tabular}[c]{@{}l@{}}Integrate/Synthesize\end{tabular} &
  \begin{tabular}[c]{@{}l@{}}Link different elements of knowledge, related representations, and \\ procedures to solve problems.\end{tabular} \\ \cline{2-3} 
 &
  Evaluate &
  Evaluate alternative problem solving strategies and solutions. \\ \cline{2-3} 
 &
  Draw conclusions &
  Make valid inferences on the basis of information and evidence. \\ \cline{2-3} 
 &
  Generalize &
  \begin{tabular}[c]{@{}l@{}}Make statements that represent relationships in more general and more \\ widely applicable terms.\end{tabular} \\ \cline{2-3} 
 &
  Justify &
  Provide mathematical arguments to support a strategy or solution. \\
\bottomrule
\end{tabular}
}
\caption{Fifteen cognitive skills and their descriptions defined in the TIMSS 2019 framework.}
\label{table:timss_skills}
\end{table*}

%% file: appendix/topics.tex
\begin{table*}[ht]
\resizebox{\textwidth}{!}{%
\begin{tabular}{lll}
\toprule
\textbf{Topics}                                                  & \textbf{Content Domain} & \textbf{Difficulty} \\ 
\midrule
1. Problems involving the digits of numbers                      & Number                  & 1, 3                \\
2. Solving for unknowns under special conditions on the solution & Algebra                 & 1, 2, 3             \\
3. Finding unknowns when two equations have the same solution    & Algebra                 & 1, 2                \\
4. Applying linear inequalities to geometric figures             & Geometry                & 2, 3                \\
5. Applying linear inequalities to pricing                       & Algebra                 & 1, 2                \\
7. Using square roots to calculate lengths in geometric figures  & Geometry                & 2, 3                \\
8. Performing simple addition and subtraction of square roots    & Number                  & 1, 2                \\
9. Applying multiplication formulas                              & Number                  & 2, 3                \\
10. Determining coefficients or constants to complete the square & Number                  & 1, 2, 3             \\
11. Solving quadratic equations by factoring                     & Algebra                 & 1, 2                \\
12. Rewriting expressions in perfect square form                 & Algebra                 & 1, 2                \\
15. Finding the quadratic function given the vertex and another point on its graph & Algebra & 1, 2 \\
\bottomrule
\end{tabular}
}
\caption{The topics and difficulty levels of problems. This information comes from the original dataset's metadata, which details the topic and difficulty level (1-3) of each problem. The isomorphic problems require the same mathematical concept to solve, but the difference in numbers makes one more tricky and complicated than the other.}
\label{table:topics}
\end{table*}

%% file: appendix/inter_rater_agreement.tex
\begin{table*}[ht]
\centering
\begin{tabular}{rccccccccccccccc}
\toprule
\textbf{Topics} & 1  & 2  & 3  & 4  & 5  & 6  & 7  & 8  & 9  & 10 & 11  & 12 & 13 & 14 & 15 \\
\textbf{\% Agreement}  & 95 & 70 & 96 & 95 & 89 & 68 & 74 & 86 & 80 & 80 & 100 & 88 & 40 & 57 & 78 \\
\bottomrule
\end{tabular}
\caption{The inter-rater percentage absolute agreement of each topic. The percentage indicates the ratio of unanimous verdicts in each teacher group.}
\label{table:inter_rater_agreement}
\end{table*}

%% file: appendix/distribution.tex
\begin{table*}[t]
\resizebox{\textwidth}{!}{%
\begin{tabular}{lcccccccc}
\toprule
 &
  \textbf{Recall} &
  \textbf{Recognize} &
  \textbf{Classify/Order} &
  \textbf{Compute} &
  \textbf{Determine} &
  \textbf{Represent} &
  \textbf{Implement} \\
  \midrule
Evident Yes        & 413 & 434 & 92 & 522 & 441 & 122 & 307 \\
Evident No         & 62  & 56  & 8  & 125 & 54  & 33  & 140 \\
Vague Yes          & 20  & 16  & 2  & 26  & 25  & 0   & 18  \\
Vague No           & 35  & 28  & 0  & 20  & 16  & 3 & 18 \\
\midrule
Student responses & 530 & 534 & 102 & 693 & 536 & 158 & 483 \\
\bottomrule
\end{tabular}
}
\caption{Distribution of verdicts and number of diagnosed student responses in each cognitive skill. Note that the number of student responses can be larger than the sum of the four labels because some diagnostic checklists have two items for the same cognitive skill.}
\label{table:distribution}
\end{table*}